\documentclass{article}

\usepackage{arxiv}
\usepackage[utf8]{inputenc}
\usepackage[T1]{fontenc}    
\usepackage{hyperref}       
\hypersetup{
    colorlinks=false,
    pdfborder={0 0 0}, 
}
\usepackage{amsmath}
\usepackage{url}            
\usepackage{booktabs}       
\usepackage{amsfonts}       
\usepackage{nicefrac}       
\usepackage{microtype}      
\usepackage{lipsum}		
\usepackage{graphicx}
\usepackage{doi}
\usepackage[square, numbers]{natbib}
\usepackage{subcaption}



\title{Non-Deterministic Finite Automata for Telugu Letters and Visualizations for Metrical Poetry}

\makeatletter
\renewcommand{\@title}{Non-Deterministic Finite Automata for Telugu Letters and Visualizations for Metrical Poetry}
\makeatother

\author{ \href{https://orcid.org/0009-0003-3233-9447}{\includegraphics[scale=0.06]{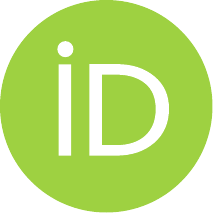}\hspace{1mm}Boddu Sri Pavan} \\
        \texttt{boddusripavan111@gmail.com} \\
	\And
	\href{https://orcid.org/0000-0001-7489-1174}{\includegraphics[scale=0.06]{orcid.pdf}\hspace{1mm}Chandrasheker Thummanagoti} \\
	\texttt{chandra.indra@rguktn.ac.in} \\
    \And
	\href{https://orcid.org/0009-0000-9389-3271}{\includegraphics[scale=0.06]{orcid.pdf}\hspace{1mm}Appana Mohan Naga Phani Kumar} \\
	\texttt{mohanappana2001@gmail.com} \\
    \And
	\href{https://orcid.org/0009-0000-3793-7743}{\includegraphics[scale=0.06]{orcid.pdf}\hspace{1mm}Boddu Swathi Sree} \\
	\texttt{n220962@rguktn.ac.in} \\
}

\date{}

\renewcommand{\headeright}{Paper}
\renewcommand{\undertitle}{Paper}
\renewcommand{\shorttitle}{Non-Deterministic Finite Automata for Telugu Letters and Visualizations for Metrical Poetry}

\hypersetup{
pdftitle={Non-Deterministic Finite Automata for Telugu Letters and Visualizations for Metrical Poetry},
pdfsubject={cs.CL},
pdfauthor={Boddu Sri Pavan, Chandrasheker Thummanagoti, Appana Mohan Naga Phani Kumar, Boddu Swathi Sree},
pdfkeywords={Non-deterministic finite automata, Telugu prosody, Computational linguistics, Syllabic weight classification, Regular expressions}
}

\begin{document}

\maketitle

\begin{abstract}

Syllabic weight is central to Telugu metrical poetry. Syllabically, every letter (aksharam token) is either light (laghuvu) or heavy (guruvu). Poets classify it using a set of pre-defined rules. This paper formalizes aksharam token, the atomic unit of syllabic classification, as a non-deterministic finite automaton (NFA) with the regular expression$\colon$~$(A_h+A_d+A_p)(\epsilon+V_s)+A_m+H^+G(\epsilon+H+V_s+V_v)$, where $\epsilon$ denotes a null string, $A$ denotes a vowel (achchu), comprising hrasvam ($A_h$), deergham ($A_d$), plutham ($A_p$), and miscellaneous ($A_m$), $H$ a pure consonant (pollu\_hallu), $G$ a vowel diacritic (gunintha\_chihnam), $V_s$ the sunna (anusvara), and $V_v$ the visarga, the latter two being nasal and aspirated sounds respectively. We then present automata diagrams and regular expressions for (i) a necessary condition satisfied by every light letter, and (ii) a sufficient condition for a letter to be heavy. Finally, we propose graphical visualizations for the 9 Telugu poem types.
\end{abstract}


\section{Introduction}
\label{sec:introduction}
\begin{figure}[t!]
\begin{center}
\includegraphics[width=0.85\textwidth]{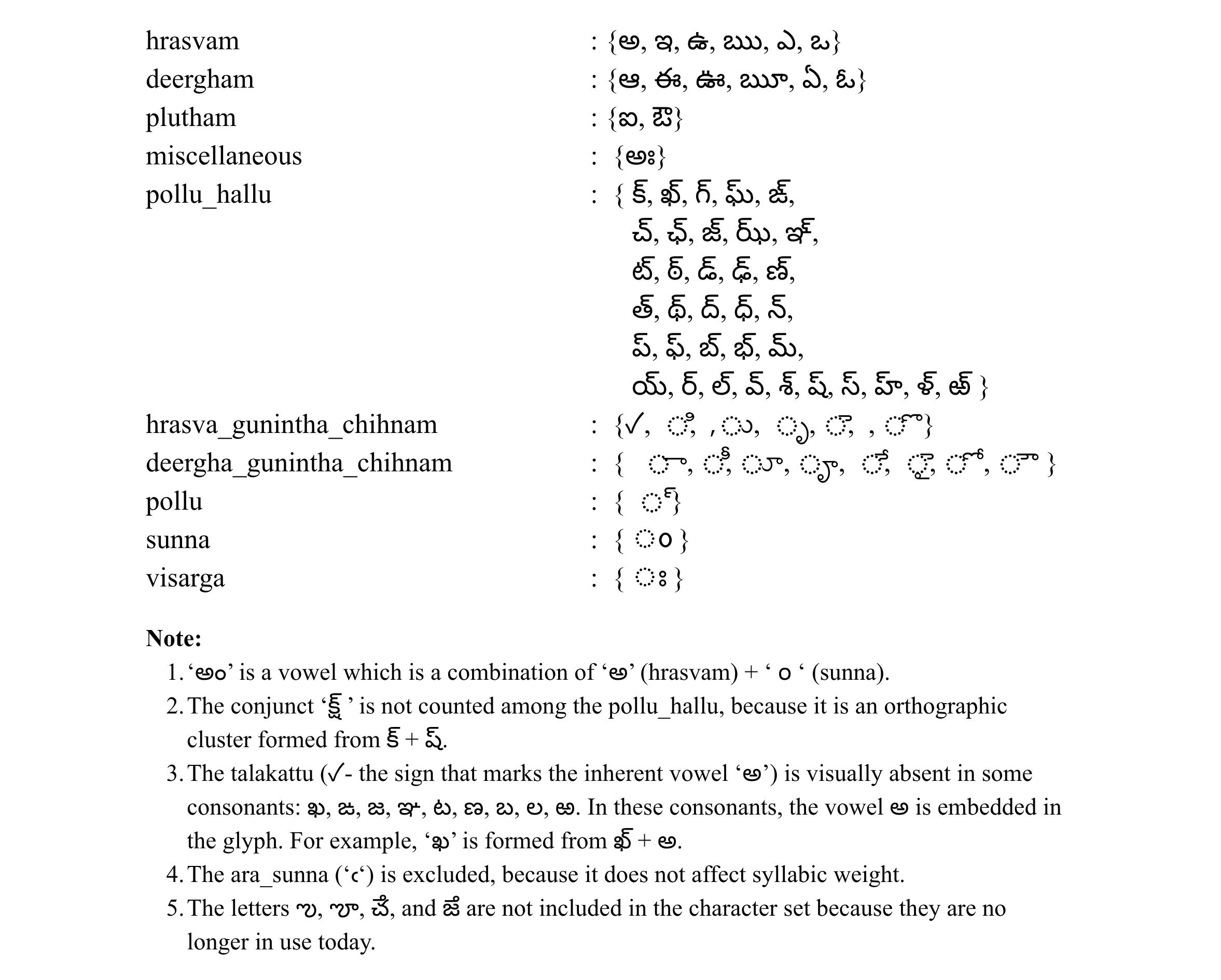}
\end{center}
\caption{Telugu character set: short vowels (hrasvam), long vowels (deergham), prolated vowels (plutham), miscellaneous, pure consonants (pollu\_hallu), short and long vowel diacritics (hrasva gunintha\_chihnam, deergha gunintha\_chihnam), pollu, sunna, and visarga.}
\label{fig:letters}
\end{figure}

\setlength{\parindent}{0pt}Telugu is a language with a literary history spanning centuries and a rich inventory of letters. We consider the character set of Telugu Vari Sampoorna Pedda Bala Siksha \cite{pedda_bala_siksha}, with the modifications shown in Figure~\ref{fig:letters}, which comprises 16 vowels (achchu- hrasvam, deergham, plutham, miscellaneous), 35 pure consonants (pollu\_hallu), the pollu, the sunna, and the visarga. The term 'Aksharam Token' was introduced in~\cite{chandassu}: it denotes a single independent unit that can be classified as light (laghuvu) or heavy (guruvu) in Telugu metrical poetry (chandassu). A character sequence is an AksharamToken if and only if it satisfies one of the following conditions:

\begin{enumerate}
  \item A vowel (achhu), whether short (hrasvam), long (deergham), prolated (plutham) or miscellaneous (described in Figure~\ref{fig:letters}).
  \item A vowel combined with a sunna. 
  \item A pure consonant (pollu\_hallu)  combined with a vowel diacritic (gunintha\_chihnam), forming a consonant. A consonant cannot carry more than one vowel diacritic at a time.
  \item Two or more pure consonants combined with a vowel diacritic, forming an orthographic cluster. A cluster cannot carry more than one vowel diacritic at a time.
  \item A consonant followed by a pure consonant (pollu\_hallu).
  \item A consonant combined with a sunna.
  \item A consonant combined with a visarga.
  \item An orthographic cluster followed by a pure consonant (pollu\_hallu).
  \item An orthographic cluster combined with a sunna.
  \item An orthographic cluster combined with a visarga.
\end{enumerate}

\section{Related Work}
\label{sec:Related Works}
The rules of Telugu prosody are found in classical treatises such as Chhan’doo vyaakarand-amu by Anjaneyamurthy \cite{chando_vyakaranamu} and Telugu Vari Sampoorna Pedda Bala Siksha \cite{pedda_bala_siksha} by Gajula Satyanarayana. Boddu Sri Pavan et al. \cite{chandassu}, released 4,651 poems (padyams) annotated with laghuvu-guruvu sequences, together with a heuristic-based aksharam tokenizer, syllabic weight generator, and a metric 'Chandassu Score' on which they report 91.73\%. Earlier tools include a chandassu-determination and word-prediction application by Raghu Korrapati et al. \cite{raghu_korrapati}, a Chandassu Recognizer by I. Reddy
Sekhar Reddy et al. \cite{reddy_sekhar_reddy}, and VerseMetrics, a Python based analyzer by S Vijay Kumar et al. \cite{nelli_ramesh}. Chandojñānam \cite{terdalkar2023chandojnanam} by Terdalkar, Hrishikesh et al. identifies metres from text or images and offers approximate and close matches for erroneous verses. For Urdu, UPMVM \cite{zaman2026upmvm} by Asia Zaman et al. develops a DFA for each identified Arud meter, matches verses against 290 meters and sub-meters, and reports an average accuracy of 94\%. These automata operate at the metre level.

Finite-state methods on Indic scripts have targeted transliteration and morphology. Tina Bögel et al. \cite{bogel2008developing} use a cascade of finite-state transducers to map Urdu and Hindi scripts into a common ASCII transcription for a shared morphological analyzer. Malik et al. \cite{malik2008hindi} apply finite-state transducers to Hindi-Urdu machine transliteration. Hellsten et al. \cite{hellsten2017transliterated} use weighted finite-state transducers for transliterated keyboard input in South Asian languages, deployed for 22 languages in Gboard. Telugu morphological analysis has followed paradigm-based by G. Uma Maheshwar Rao et al. \cite{rao2011telugu} and machine learning based approaches by Sai Kiranmai et al. \cite{kiranmai2010morphological}, both at the word level.

A non-deterministic finite automata (NFA), introduced by M. O. Rabin and D. Scott in 1959\cite{rabin1959finite}, is a 5-tuple $(Q, \Sigma, \delta, q_0, F)$. Here $Q$ is a finite, non-empty set of states, $\Sigma$ is a finite alphabet, $\delta : Q \times (\Sigma \cup \{\epsilon\}) \to \mathcal{P}(Q)$ is the transition function, $q_0 \in Q$ is the start state, and $F \subseteq Q$ is the set of accepting states (final states).

Rodolfo Delmonte's SPARSAR visualization \cite{delmonte2015visualizing} converts the system's analysis of a poem into ten graphical maps, grouped into phonetic, poetic (rhyme and metrical structure), and semantic views, and uses colour to highlight contrasts between sound classes and word polarities. A. Abdul-Rahman et al. \cite{abdulrahman2013rule} present a user-centered design study that uses rule-based visual mappings to manage a large set of poetic variables while reducing the effort of operating the mapping control panel, and they compare design options for visualizing poems latitudinally. To our knowledge, neither work targets Telugu chandassu, so neither encodes laghuvu and guruvu weights, ganam types, or yati and prasa positions, which our padyam plots show in a single view.

Prior Telugu tools compute syllabic weight in code, prior automata for prosody operate after weights have been assigned. This paper makes four contributions:
\begin{enumerate}
    \item \textbf{High-level NFA}: A non-deterministic finite automaton for the aksharam token, with its regular expression.
    \item \textbf{Prosody-aware NFA}: A refined automaton that separates short and long vowels and diacritics, with its regular expression and a regular expression for the orthographic cluster.
    \item \textbf{Weight conditions}: Two conditions derived from the prosody-aware NFA: a necessary condition satisfied by every light token (laghuvu) and a sufficient condition for a heavy token (guruvu). 
    \item \textbf{Padyam plots}: A graphical visualization of Telugu poem types that encodes syllabic weight, yati, prasa, and ganam in a single view, making their metrical structure easier to read.
\end{enumerate}

\section{Proposed Methodology for NFA}
\label{sec:methodology}

\begin{figure}[t!]
\begin{center}
\includegraphics[width=0.85\textwidth]{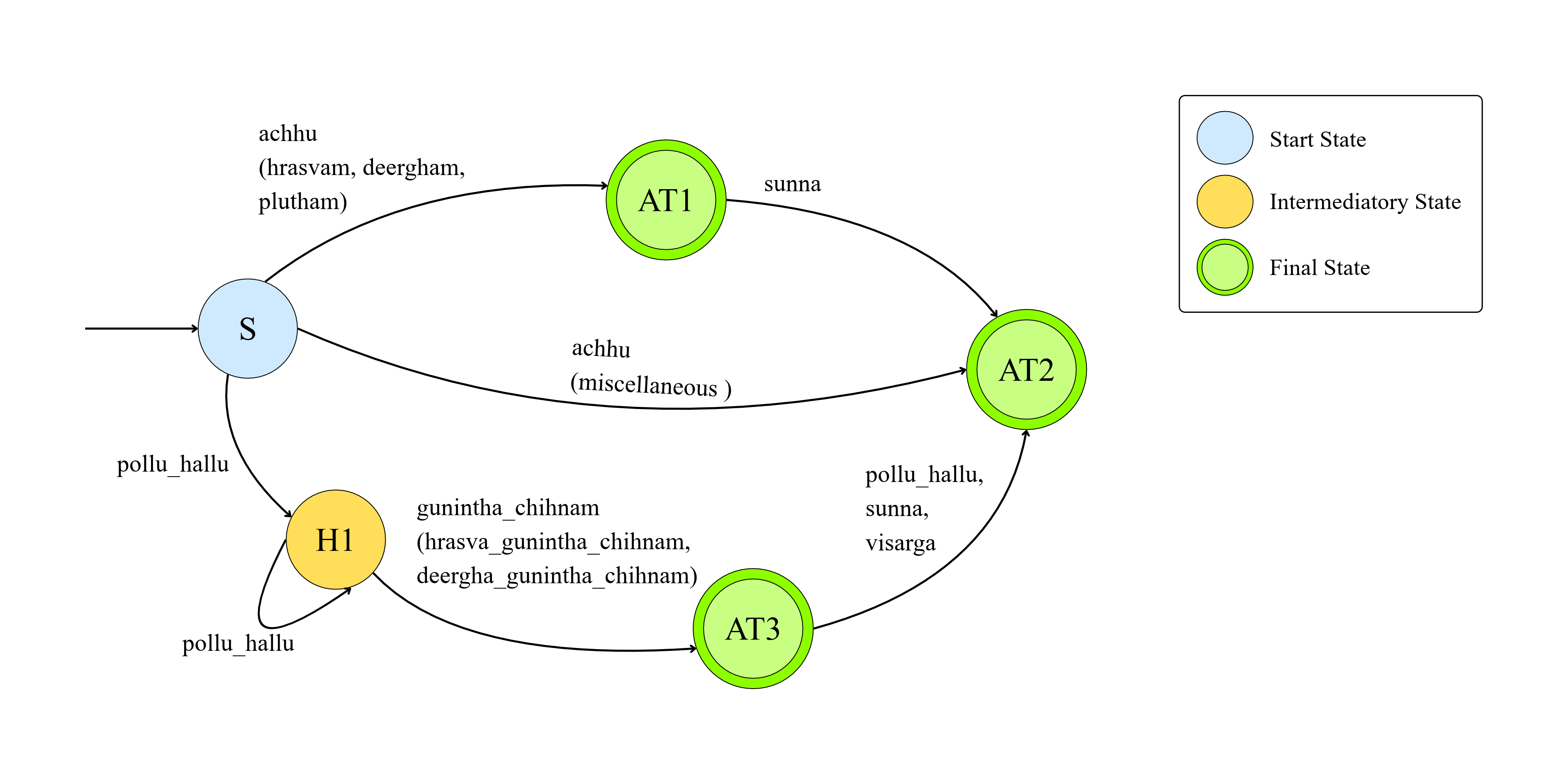}
\end{center}
\caption{NFA for Aksharam Token}
\label{fig:aksharam_token_high}
\end{figure}

Aggregating the rules in the Introduction \ref{sec:introduction}, we propose high-level non-deterministic finite automata (NFA) for the aksharam token, shown in Figure \ref{fig:aksharam_token_high}. Let $A$ denotes achhu or vowels, comprising hrasvam ($A_h$), deergham ($A_d$), plutham ($A_p$), and miscellaneous ($A_m$). $H$ the pollu\_hallu, $G$ the gunintha\_chihnam, comprising hrasva\_gunintha\_chihnam ($G_s$), and deergha\_gunintha\_chihnam ($G_d$), $V_s$ the sunna, $V_v$ the visarga. Each is a non-empty set of input symbols. The 5-tuple of the NFA is:
\begin{enumerate}
    \item $Q$= \{S, H1, AT1, AT2, AT3\}
    \item $q_{0}$= \{S\}, marked by '$\rightarrow$'
    \item $\sum$= $A$ $\cup$ $A_m$ $\cup$ $H$ $\cup$ $G$ $\cup$ $V_s$ $\cup$ $V_v$ (Here, $A_m$ is included separately as it cannot be combined with sunna-$V_s$)
    \item $\delta$: Table \ref{tab:aksharam_token_high}
    \item $F$= \{AT1, AT2, AT3\}, marked by '*'
\end{enumerate}

\begin{table}[b!]
\centering
\begin{tabular}{lllllll}
\toprule
& $A$ & $A_m$& $H$& $G$& $V_s$& $V_v$ \\
\midrule
$\rightarrow$S& AT1* & AT2*& H1& $\phi$& $\phi$& $\phi$\\
H1& $\phi$& $\phi$& H1& AT3*& $\phi$& $\phi$\\
AT1& $\phi$& $\phi$& $\phi$& $\phi$& AT2*& $\phi$\\
AT3& $\phi$& $\phi$& AT2*& $\phi$& AT2*& AT2*\\
\bottomrule
\end{tabular}
\vspace{0.6em}
\caption{NFA: Transition function table for Aksharam Token (where $\phi$ is empty set)}
\label{tab:aksharam_token_high}
\end{table}

The regular expression of the NFA is: $(A_h+A_d+A_p)(\epsilon+V_s)+A_m+H^+G(\epsilon+H+V_s+V_v)$

\begin{figure}[t!]
\begin{center}
\includegraphics[width=0.85\textwidth]{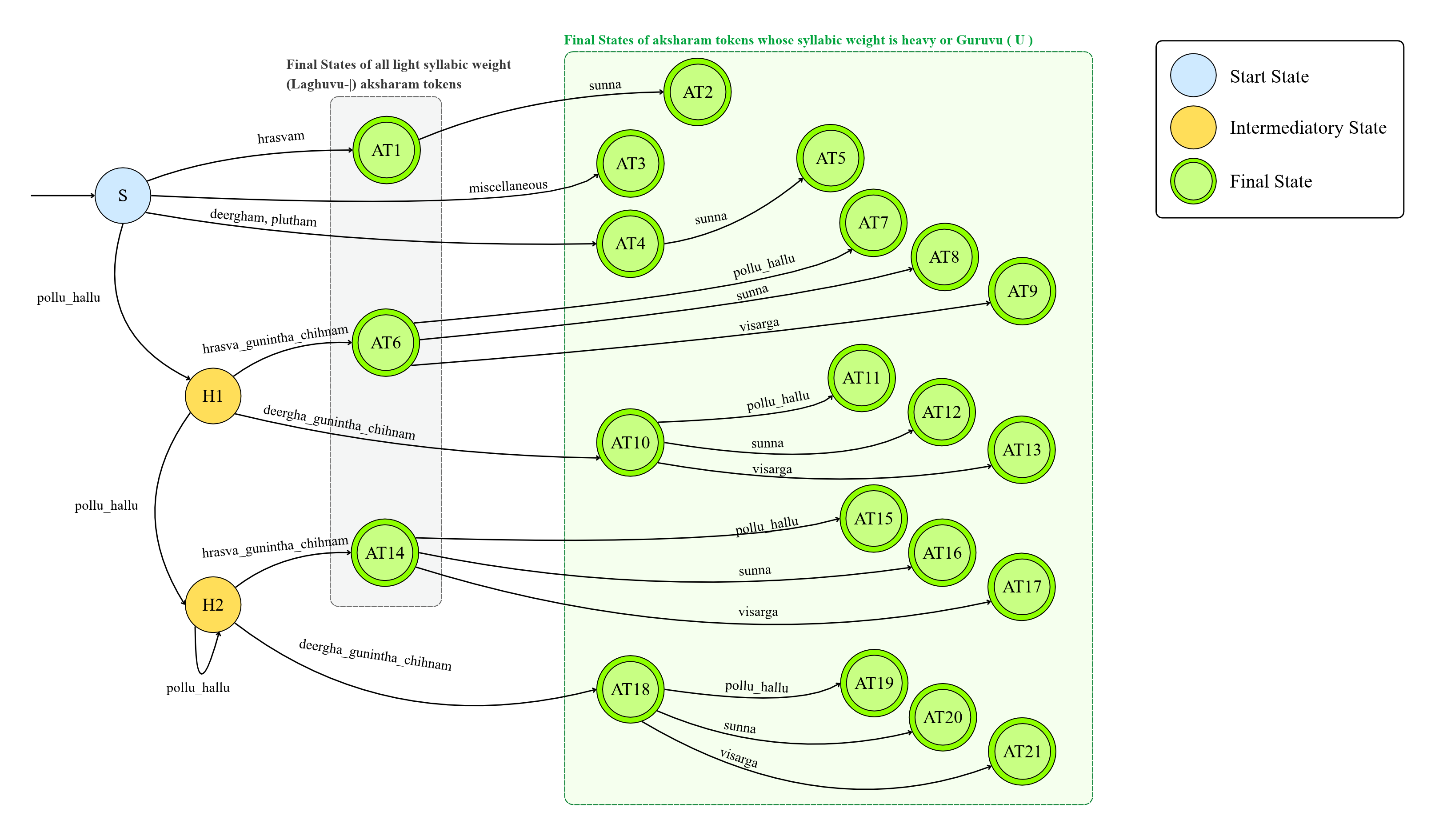}
\end{center}
\caption{Prosody-aware NFA for Aksharam Token}
\label{fig:aksharam_granular_level_architecture}
\end{figure}

This NFA accepts aksharam tokens, the atomic units of metrical poetry, each of which is classified as light (laghuvu) or heavy (guruvu). Because vowels include both short and long forms, the transition from 'S' to 'AT1' can be split further. Likewise, because vowel diacritics have short and long forms, the transition from 'H1' to 'AT3' can be split. The resulting syllabic-weight-aware automaton is shown in Figure~\ref{fig:aksharam_granular_level_architecture}. Its 5-tuple is:

\begin{enumerate}
    \item $Q$= \{S, H1, H2, AT1, AT2, AT3, AT4, AT5, AT6, AT7, AT8, AT9, AT10, AT11, AT12, AT13, AT14, AT15, AT16, AT17, AT18, AT19, AT20, AT21\}
    \item $q_{0}$= \{S\}; marked by '$\rightarrow$'
    \item $\sum$= $A_h$ $\cup$ $A_d$ $\cup$ $A_p$ $\cup$ $A_m$ $\cup$ $H$ $\cup$ $G_h$ $\cup$ $G_d$ $\cup$ $V_s$ $\cup$ $V_v$
    \item $\delta$: Table \ref{tab:aksharam_token_granular}
    \item $F$= \{AT1, AT2, AT3, AT4, AT5, AT6, AT7, AT8, AT9, AT10, AT11, AT12, AT13, AT14, AT15, AT16, AT17, AT18, AT19, AT20, AT21\}; marked by '*'
\end{enumerate}

\begin{table}[b!]
\centering
\begin{tabular}{lllllllllll}
\toprule
& $A_h$ & $A_d$& $A_p$& $A_m$& $H$& $G_h$& $G_d$& $V_s$& $V_v$& \\
\midrule
$\rightarrow$S& AT1* & AT4*& AT4*& AT3*& H1& $\phi$& $\phi$& $\phi$& $\phi$&\\
AT1& $\phi$& $\phi$& $\phi$& $\phi$& $\phi$& $\phi$& $\phi$& AT2*& $\phi$&\\
AT4& $\phi$& $\phi$& $\phi$& $\phi$& $\phi$& $\phi$& $\phi$& AT5*& $\phi$&\\
H1& $\phi$& $\phi$& $\phi$& $\phi$& H2& AT6*& AT10*& $\phi$& $\phi$&\\
AT6& $\phi$& $\phi$& $\phi$& $\phi$& AT7*& $\phi$& $\phi$& AT8*& AT9*&\\
AT10& $\phi$& $\phi$& $\phi$& $\phi$& AT11*& $\phi$& $\phi$& AT12*& AT13*&\\
H2& $\phi$& $\phi$& $\phi$& $\phi$& H2& AT14*& AT18*& $\phi$& $\phi$&\\
AT14& $\phi$& $\phi$& $\phi$& $\phi$& AT15*& $\phi$& $\phi$& AT16*& AT17*&\\
AT18& $\phi$& $\phi$& $\phi$& $\phi$& AT19*& $\phi$& $\phi$& AT20*& AT21*&\\
\bottomrule
\end{tabular}
\vspace{0.6em}
\caption{Prosody-aware NFA: Transition table for Aksharam Token (where $\phi$ is empty set)}
\label{tab:aksharam_token_granular}
\end{table}

States AT1, AT2, AT3, AT4, AT5 accept a bare vowel or a vowel followed by a sunna. The remaining final states accept tokens that contain pure consonants: AT6 to AT13 accept tokens without, and AT14 to AT21 accept tokens with orthographic cluster. The regular expression of this syllabic weight-aware NFA is: $(A_h+A_d+A_p)(\epsilon+V_s)+A_m+H^{+}(G_h+G_d)(\epsilon+H+V_s+V_v)$. The regular expression of an orthographic cluster is: $HH^+(G_h+G_d)(\epsilon+H+V_s+V_v)$

From the syllabic weight-aware NFA in Figure~\ref{fig:aksharam_granular_level_architecture}, we derive two propositions:

\begin{enumerate}
    \item \textbf{Proposition 1 (necessary condition for laghuvu):} Every light syllabic weight (laghuvu-|) aksharam token has the form $A_h+H^+G_h$.
    \item \textbf{Proposition 2 (sufficient condition for guruvu):} Every aksharam token of the form $A_hV_s+(A_d+A_p)(\epsilon+V_s)+A_m+H^+((G_h+G_d)(H+V_s+V_v)+G_d)$ is heavy (guruvu-U).
\end{enumerate}

As described in the classical texts \cite{chando_vyakaranamu}, \cite{pedda_bala_siksha}, the weight of an aksharam token also depends on the token that follows it. If the following token is an orthographic cluster, the current token is heavy, regardless of its intrinsic weight. Proposition 1 is derived from the final states AT1, AT6, and AT14 (light grey in Figure~\ref{fig:aksharam_granular_level_architecture}). Proposition 2 is derived from all other final states (light green).

\section{PadyamPlot Visualization}
\label{sec:padyamplot}

We propose a visualization plot for Telugu metrical poetry (padyam), in which each cell represents one aksharam token and its color conveys syllabic weight, position, and ganam type (sequence of syllables). The paadams (lines) of the poem run from top to bottom. Within each line, the ganams (syllabic sequences) run from left to right. Each ganam (syllabic sequence) is drawn as a vertical block whose cells hold the syllabic weights of its aksharam tokens, read from bottom to top.

Weight and position (cell fill and hatching):
\begin{enumerate}
\item An empty cell denotes a light token (laghuvu, |).
\item A shaded cell (light green) denotes a heavy token (guruvu, U). 
\item Vertical hatching marks a prasa position. It is applied to an empty cell for a light token and to a shaded cell for a heavy token.
\item Horizontal hatching marks a yati position, applied in the same way.
\item Inclined hatching marks a position that is prasa yati, applied in the same way.
\end{enumerate}

Ganam type (cell border):
\begin{enumerate}
\item A thick border denotes a Surya ganam.
\item A dashed border denotes an Indra ganam.
\end{enumerate}

\begin{figure}[t!]
\centering
\begin{subfigure}[b]{0.33\textwidth}
  \centering
  \includegraphics[width=\linewidth]{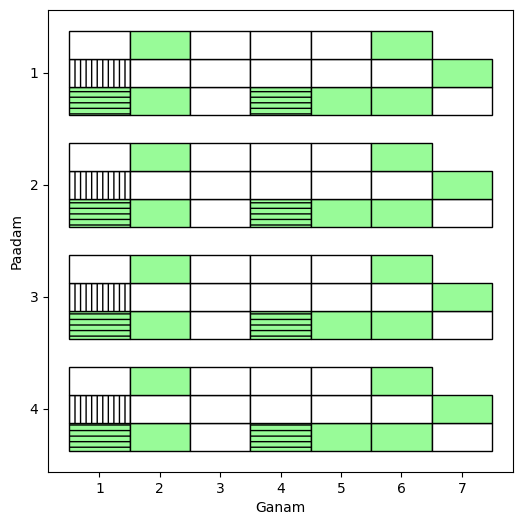}
  \caption{\emph{Vutpalamaala}}
  \label{fig:vutpalamaala}
\end{subfigure}%
\hfill
\begin{subfigure}[b]{0.33\textwidth}
  \centering
  \includegraphics[width=\linewidth]{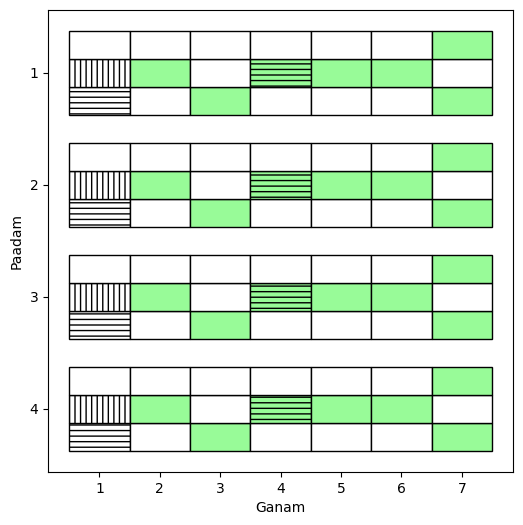}
  \caption{\emph{Champakamala}}
  \label{fig:champakamala}
\end{subfigure}%
\hfill
\begin{subfigure}[b]{0.33\textwidth}
  \centering
  \includegraphics[width=\linewidth]{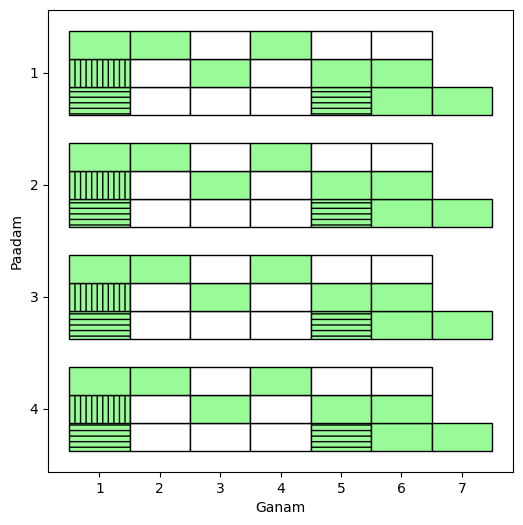}
  \caption{\emph{Saardulamu}}
  \label{fig:sardulamu}
\end{subfigure}

 \vspace{0.6cm}
 
\centering
\begin{subfigure}[b]{0.33\textwidth}
  \centering
  \includegraphics[width=\linewidth]{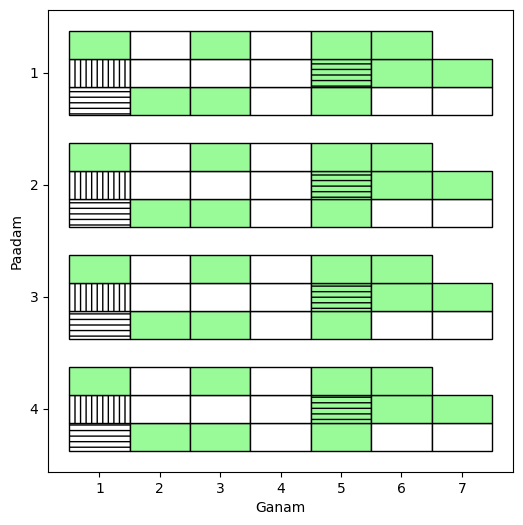}
  \caption{\emph{Mattebhamu}}
  \label{fig:mattebhamu}
\end{subfigure}%
\hfill
\begin{subfigure}[b]{0.33\textwidth}
  \centering
  \includegraphics[width=\linewidth]{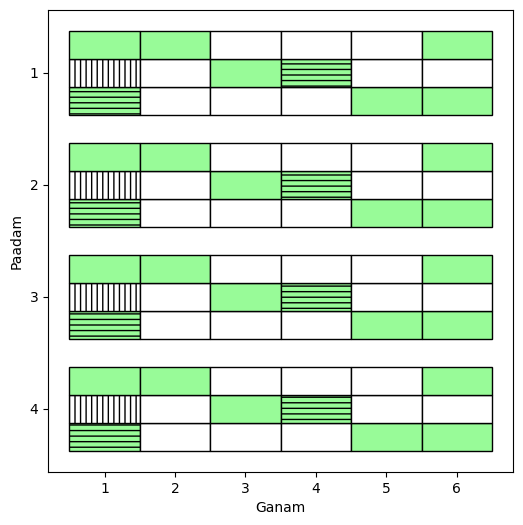}
  \caption{\emph{Mattakokila}}
  \label{fig:mattakokila}
\end{subfigure}%
\hfill
\begin{subfigure}[b]{0.33\textwidth}
  \centering
  \includegraphics[width=\linewidth]{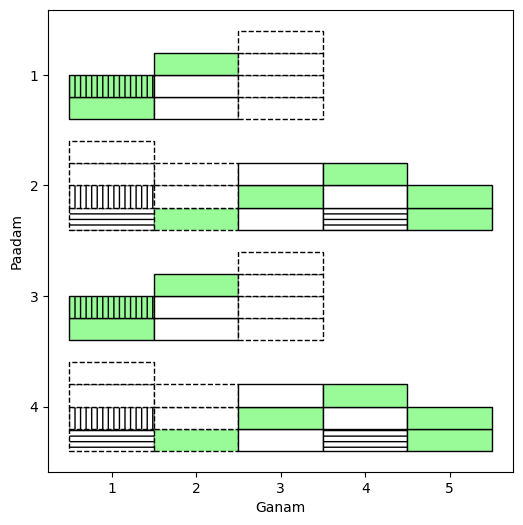}
  \caption{\emph{Kandamu}}
  \label{fig:kandamu}
\end{subfigure}

 \vspace{0.6cm}
 
\centering
\begin{subfigure}[b]{0.33\textwidth}
  \centering
  \includegraphics[width=\linewidth]{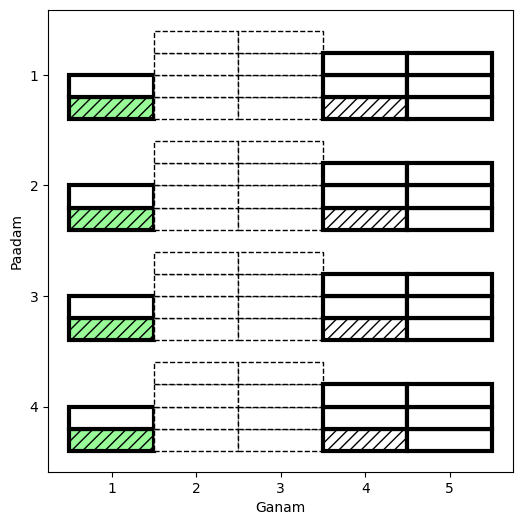}
  \caption{\emph{Teytageethi}}
  \label{fig:teytageethi}
\end{subfigure}%
\hfill
\begin{subfigure}[b]{0.33\textwidth}
  \centering
  \includegraphics[width=\linewidth]{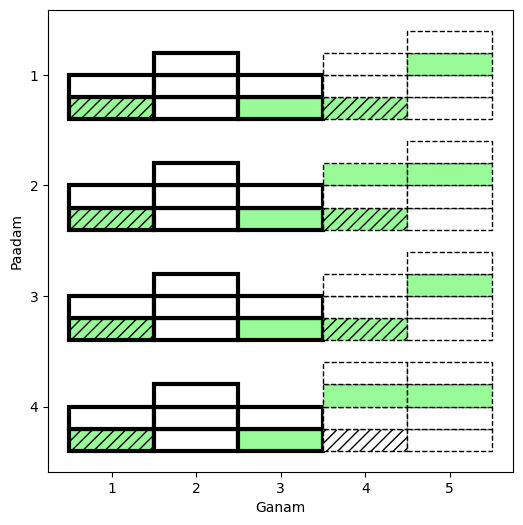}
  \caption{\emph{Aataveladi}}
  \label{fig:aataveladi}
\end{subfigure}%
\hfill
\begin{subfigure}[b]{0.33\textwidth}
  \centering
  \includegraphics[width=\linewidth]{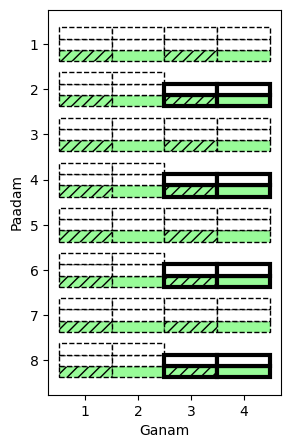}
  \caption{\emph{Seesamu}}
  \label{fig:seesamu}
\end{subfigure}

\caption{PadyamPlot Visualization}
\label{fig:padyamplots}
\end{figure}

In Figure~\ref{fig:padyamplots}, we present these plots for the following 9 poem types: Vutpalamaala, Champakamaala, Saardulamu, Mattebhamu, Mattakokila, Kandamu, Teytageethi, Aataveladi, and Seesamu, with the dataset mentioned in the Appendix \ref{sec:appendix}.


\section{Conclusion}
We presented non-deterministic finite automata (NFAs) and regular expressions for the aksharam token at two levels: high-level and syllabic-weight-aware. We formulated a regular expression for the orthographic cluster. From the weight-aware automaton, we derived two propositions: a necessary condition satisfied by every light token (laghuvu) and a sufficient condition for an aksharam token to be heavy (guruvu). We also proposed padyam plots, graphical visualizations of Telugu metrical poems that encode syllabic weight, yati, prasa, and ganam, etc. We expect these formal models and visualizations to support further computational work on Telugu chandassu.

\section{Acknowledgements}
We thank Kamatham Tharun Kumar for proofreading this paper. We are grateful to our parents and family members for their support. This work is completed by the grace of Sri Ramalinga Chowdeswari Devi.

\bibliographystyle{plain}

\section{Appendix}
\label{sec:appendix}

For this work, we considered ganams upto length of four aksharam tokens as shown in Table \ref{tab:cat_ganam_at}, with two functional classes: INDRA Ganams (BHA, RA, THA, NALA, NAGA, SALA) and SURYA Ganams (HA, NA).
\begin{table}[b!]
\centering
\begin{tabular}{lll}
\toprule
Aksharam Token Count& Ganam Name& LaghuvuGuruvu Pattern\\
\midrule
1& LA& | \\
1& GA& U \\
2& LAA& || \\
2& VA& |U \\
2& HA& U| \\
2& GAA& UU \\
3& NA& ||| \\
3& SA& ||U \\
3& JA& |U| \\
3& YA& |UU \\
3& BHA& U|| \\
3& RA& U|U \\
3& THA& UU| \\
3& MA& UUU \\
4& NALA& |||| \\
4& NAGA& |||U \\
4& SALA& ||U| \\
\bottomrule
\end{tabular}
\vspace{0.6em}
\caption{Ganam classification by aksharam token count and laghuvu-guruvu pattern}
\label{tab:cat_ganam_at}
\end{table}

Every padyam type below consists of four paadalu (lines).

\begin{enumerate}
    \item \textbf{Vutpalamaala:} Each paadam has 20 aksharam tokens and the ganam sequence BHA-RA-NA-BHA-BHA-RA-VA. The yati position is the 10th aksharam token. Prasa applies.

    \item \textbf{Champakamaala:} Each paadam has 21 aksharam tokens and the ganam sequence NA-JA-BHA-JA-JA-JA-RA. The yati position is the 11th aksharam token. Prasa applies.

    \item \textbf{Saardulamu:} Each paadam has 19 aksharam tokens and the ganam sequence MA-SA-JA-SA-THA-THA-GA. The yati position is the 13th aksharam token. Prasa applies.

    \item \textbf{Mattebhamu} Each paadam has 20 aksharam tokens and the ganam sequence SA-BHA-RA-NA-MA-YA-VA. The yati position is the 14th aksharam token. Prasa applies.

    \item \textbf{Mattakokila} Each paadam has 18 aksharam tokens and the ganam sequence RA-SA-JA-JA-BHA-RA. The yati position is the 11th aksharam token. Prasa applies.

    \item \textbf{Kandamu} Each paadam contains the ganams from: GAA, BHA, JA, SA, and NALA. The ganams at odd positions (including the first) must be GAA, BHA, SA, or NALA. The last ganam of the second and fourth paadalu must be GAA or SA, and the 3rd ganam of these paadalu must be JA or NALA. Yati applies only to the second and fourth paadalu, at the first aksharam token of the 4th ganam. Prasa applies.

    \item \textbf{Teytageethi:} Each paadam has the ganam sequence SURYA-INDRA-INDRA-SURYA-SURYA. The yati position is the first aksharam token of the 4th ganam of each paadam. In place of yati, the variant prasa yati applies.
    
    \item \textbf{Aataveladi:} Each paadam has the ganam sequence SURYA-SURYA-SURYA-INDRA-INDRA. The yati position is the first aksharam token of the 4th ganam of each paadam. In place of yati, the variant prasa yati applies.

    \item \textbf{Seesamu:} Each paadam has the ganam sequence INDRA-INDRA-INDRA-INDRA-INDRA-INDRA-SURYA-SURYA and is split into two sub-paadalu of four ganams each. The first sub-paadam contains four INDRA ganams, and the second contains INDRA-INDRA-SURYA-SURYA. The yati position is the first aksharam token of the 3rd ganam of each sub-paadam. In place of yati, the variant prasa yati applies. Prasa is optional.
\end{enumerate}

\end{document}